\def\year{2018}\relax
\documentclass[letterpaper]{article} %DO NOT CHANGE THIS
\usepackage{aaai18}  %Required
\usepackage{times}  %Required
\usepackage{helvet}  %Required
\usepackage{courier}  %Required
\usepackage{url}  %Required
\usepackage{graphicx}  %Required
\usepackage{multirow}
\usepackage{pgfplotstable}
\usepackage{capt-of}
\usepackage{amsmath}
\usepackage{amssymb}

\begin{document}
\title{Generative Adversarial Network for Abstractive Text Summarization\thanks{The work was partially supported by CAS Pioneer Hundred Talents Program and the MOE Key Laboratory of Machine Perception at Peking University under grant number K-2017-02. Q. Qu is the corresponding author.}}
\author{Linqing Liu$^{1}$ \hspace{0.3cm}  Yao Lu$^2$ \hspace{0.3cm} Min Yang$^1$ \hspace{0.3cm} Qiang Qu$^{1,4}$ \hspace{0.3cm} Jia Zhu$^3$ \hspace{0.3cm} Hongyan Li$^4$\\
	$^1$Shenzhen Institutes of Advanced Technology, Chinese Academy of Sciences\\ 
	$^2$Alberta Machine Intelligence Institute \space
	$^3$School of Computer Science, South China Normal University\\ 
    $^4$MOE Key Laboratory of Machine Perception, Peking University\\ 
	%{\tt likicode@gmail.com}
	{\tt \{likicode, 95luyao, min.yang1129\}@gmail.com} \hspace{0.3cm} {\tt qiang@siat.ac.cn} \\{\tt jzhu@m.scnu.edu.cn} \hspace{0.3cm} {\tt lihy@cis.pku.edu.cn}
}

\maketitle
\begin{abstract}
In this paper, we propose an adversarial process for abstractive text summarization, in which we simultaneously train a generative model $G$ and  a discriminative model $D$.  In particular,  we build the generator $G$ as an agent of reinforcement learning, which takes the raw text as input and predicts the abstractive summarization.  We also build a discriminator which attempts to distinguish the generated summary from the ground truth summary.  Extensive experiments demonstrate that our model achieves competitive ROUGE scores with the state-of-the-art methods on CNN/Daily Mail dataset.  Qualitatively, we show that our model is able to generate more abstractive, readable and diverse summaries\footnote{Supplemental material: http://likicode.com/textsum/}. 
\end{abstract}

\section{Introduction}
Abstractive text summarization is the task of generating a short and concise summary that captures the salient ideas of the source text.  The generated summaries potentially contain new phrases and sentences that may not appear in the source text. 
In the past decades, a flurry of studies have been conducted on abstractive text summarization \cite{nallapati2016abstractive,see2017get,paulus2017deep}. Despite the remarkable progress of previous studies, abstractive summarization is still challenged by (i) Neural sequence-to-sequence models tend to generate trivial and generic summary, often involving high-frequency phrases; (ii) The generated summaries have limited grammaticality and readability; (iii) In most previous work the standard sequence-to-sequence models are trained to predict the next word in summary using the maximum-likelihood estimation (MLE) objective function. However, this strategy has two major shortcomings. First, the evaluation metric is different from the training loss. Second, the input of the decoder in each time step is often from the true summary during the training. Nevertheless, in the testing phase, the input of the next time step is the previous word generated by the decoder. This exposure bias leads to error accumulation at test time.

To address the above challenge, in this paper, we propose an adversarial framework to jointly train a generative model $G$ and a discriminative model $D$.  Specifically, the generator $G$ takes the original text as input and generate the summary. We use reinforcement learning (i.e., policy gradient) to optimize $G$ for a highly rewarded summary. Thus, it effectively bypasses exposure bias and non-differentiable task metrics issues.  We implement the discriminator $D$ as a text classifier that learns to classify the generated summaries as machine or human generated. The generator $G$ and the discriminator $D$ are optimized with a minimax two-player game. The discriminator $D$ tries to distinguish the ground truth summaries from the generated summaries by the generator $G$, while the training procedure of generator $G$ is to maximize the probability of $D$ making a mistake. Thus, this adversarial process can eventually adjust $G$ to generate plausible and high-quality abstractive summaries.

\section{Our model}
Similar to the standard training strategy \cite{goodfellow2014generative}, we simultaneously train two models in an adversarial manner: a generative model $G$ and a discriminative model $D$.  We first pre-train the generative model by generating summaries given the source text. Then we pre-train the discriminator by providing positive examples from the human-generated summaries and the negative examples produced from the pre-trained generator. After the pre-training, the generator and discriminator are trained alternatively.

\subsection{Generative Model}
The generator takes the source text $x=\{w_{1}, w_{2}, ..., w_{n}\}$ as input and predicts the summary $\hat{y} =\{\hat{y_{1}}, \hat{y_{2}}, ..., \hat{y_{m}}\}$. Here, the $n$ is the length of the source text $x$ and $m$ is the length of the predicted summary. We use a bi-directional LSTM encoder to convert the input text $x$ into a sequence of hidden states $h=\{h_{1},\dots,h_{n}\}$.  Following \cite{see2017get}, on time step $t$, an attention-based LSTM decoder is then used to compute the hidden state $s_t$ of the decoder and a context vector $c_t$.  The reader can refer to the supplement of this paper (or  \cite{see2017get}) for the implementation details. The parameters of the generator $G$ are collectively represented by $\theta$.  The context vector $c_t$ is concatenated with the decoder state $s_t$ and fed through a fully connected layer and a softmax layer to produce the  probability of predicting word from target vocabulary at each time step $t$:
{\footnotesize

	\begin{equation}
	{ P_{vocab}(\hat{y_t}) = softmax(V^{'}(V[s_{t}, c_{t}] + b) + b^{'}) } \nonumber
	\end{equation}
    
}
where $V^{'}$,  $V$, $b$, $b^{'}$ are learnable parameters.  Similar to the work of \cite{see2017get}, we incorporate a switching pointer-generator network to use either word generator from fixed vocabulary or pointer copying rare or unseen from the input sequence. Finally, we can get the final probability $P(\hat{y_t})$ of each token $\hat{y_t}$ in the summary. 

\subsection{Discriminative Model}

The discriminator is a binary classifier and aims at distinguishing the input sequence as originally generated by humans or synthesized by machines. We encode the input sequence with a CNN as it shows great effectiveness in text classification \cite{kim2014convolutional}. We use multiple filters with varying window sizes to obtain different features and then apply a max-over-time pooling operation over the features. These pooled features are passed to a fully connected softmax layer whose output is the probability of being “original”. 

\subsection{Updating model parameters}
In the adversarial process, using the discriminator as a reward function can further improve the generator iteratively by dynamically updating the discriminator. Once we obtain more realistic and high-quality summaries generated by generator $G$, we re-train the discriminator as:  
{\scriptsize
	\begin{equation}
	\min_{\phi} -\mathbf{E}_{Y\sim p_{data}}[log D_{\phi}(Y)] - \mathbf{E}_{Y\sim G_{\theta}}[log(1- D_{\phi}(Y))] \nonumber
	\end{equation}
}
When the discriminator $D$ is obtained and fixed, we are ready to update the generator $G$. The loss function of our generator $G$ consists two parts: the loss computed by policy gradient (denoted by $J_{pg}$) and the maximum-likelihood loss (denoted by $J_{ml}$). 
Formally, the objective function of $G$ is $J=\beta J_{pg} + (1-\beta) J_{ml}$, where $\beta$ is the scaling factor to balance the magnitude difference between $J_{pg}$ and $J_{ml}$.  According to the policy gradient theorem \cite{sutton2000policy},  we compute the gradient of $J_{pg}$  w.r.t. the parameters $\theta$:

{\scriptsize
	\begin{align}
	\triangledown_{\theta} J_{pg}= \frac{1}{T} \sum_{t=1}^{T} \sum_{y_{t}} R_{D}^{G_{\theta}}((Y_{1:t-1}, X), y_{t}) \cdot \triangledown_{\theta}(G_{\theta}(y_{t}|Y_{1:t-1}, X)) \nonumber\\
	= \frac{1}{T} \sum_{t=1}^{T} \mathbf{E}_{y_{t}\in G_{\theta} } [R_{D}^{G_{\theta}}((Y_{1:t-1}, X), y_{t}) 
	\triangledown_{\theta} \log p(y_{t}|Y_{1:t-1}, X)] \nonumber
	\end{align}
}
where $R_{D}^{G_{\theta}}((Y_{1:t-1}, X), y_{t})$ is the action-value function, and we have $R_{D}^{G_{\theta}}((Y_{1:t-1}, X), y_{t})={D_{\phi}}(Y_{1:T})$, $T$ is the length of the text. We update the parameters using stochastic gradient descent, $Y_{1:t}$ is the generated summary up to time step $t$,  $X$ is the source text to be condensed. 

\section{Experiments}

\textbf{Dataset - CNN/Daily Mail Corpus.} The dataset \cite{nallapati2016abstractive} is widely used in abstractive summarization. It comprises news stories in CNN and Daily Mail websites paired with multi-sentence human generated abstractive summaries. It contains 287,226 training pairs, 13,368 validation pairs and 11,490 test pairs. 
%There are 781 tokens on average of articles and 56 tokens of summaries.

\begin{table}[]
	\footnotesize
	\centering
	\begin{tabular}{|c|c|c|c|c|c|}
		\hline
		Methods & ROUGE-1 & ROUGE-2 & ROUGE-L & Human\\ \hline
		ABS &   35.46   & 13.30 & 32.65  & 2.07\\ \hline
		PGC &   39.53   & 17.28 &  36.38 & 3.81\\ \hline
		DeepRL  &   39.87   & 15.82 &  \textbf{36.90}&3.04\\ \hline
		Pretrain & 38.82 & 16.81 & 35.71 & 3.70 \\ \hline
		Ours   &   \textbf{39.92} & \textbf{17.65} & 36.71 & \textbf{4.01} \\ \hline
	\end{tabular}
	\caption{Quantitative evaluation results}
	\label{my-label}
\end{table}

\textbf{Experimental Results.} We compare our approach with three methods, including the abstractive model (ABS) \cite{nallapati2016abstractive}, the pointer-generator coverage networks (PGC) \cite{see2017get}, and the abstractive deep reinforced model (DeepRL) \cite{paulus2017deep} (ML+RL version).

We firstly compare our model with the pre-trained generator. After adversarial training, ROUGE-1, ROUGE-2, ROUGE-L increase by 1.10, 0.84 and 1.00 absolute points respectively. In addition, our model exhibits competitive ROUGE scores with the state-of-the-art methods. Specifically, our approach achieves the best ROUGE-1 and ROUGE-2 scores. 

We also perform human evaluation to evaluate the readability and quality of summaries. We randomly select 50 test examples from the dataset. For each example, two human evaluators are asked to rank each summary generated by all 5 models based on their readability, where 1 indicates the lowest level of readability while 5 indicates the highest level. As we can observe from Table 1,  our model contributes significantly to improving the readability of summaries.

To evaluate the proposed model qualitatively, we also report the generated summaries in supplementary files.

\section{Conclusion}
In this paper, we proposed an adversarial process for abstractive text summarization. Experimental results showed that our model could generate more abstractive, readable and diverse summaries.

\bibliography{sample}
\bibliographystyle{aaai}

\end{document}